\documentclass{hld2025}

\usepackage{tikz-cd}
\usepackage{multicol}
\usepackage{booktabs}

\title[Orthogonal Gradient Constraints]{Orthogonal Gradient Constraints Shape Noisy-Label Memorization Dynamics}
\hldauthor{%
\Name{Richard Mai} \Email{ricardooorichado@gmail.com}\\
\addr {Independent Researcher} }
\date{\today}

\begin{document}
\maketitle

\begin{abstract}
Modern neural networks can fit corrupted training labels, making noisy-label learning a useful setting for studying memorization-driven overfitting. Most regularization methods modify the objective, architecture, or data distribution; here we instead study a geometric intervention on the optimizer update itself. We evaluate OrthoGrad, which removes the component of each weight gradient parallel to the current weight vector, in noisy-label image classification. On MNIST with small-data regimes, OrthoGrad improves test accuracy most clearly for CNNs while reducing corrupted-label fitting. Mechanism diagnostics based on weight norms and gradient-weight cosine similarity suggest that the projection has the strongest effect when the raw gradient contains a nontrivial radial component, and becomes weaker in larger-data regimes where gradients are already nearly orthogonal to weights. Additional CIFAR-10 ResNet-18 experiments show that the method can alter memorization trajectories but does not prevent eventual noisy-label memorization. These results support orthogonal update constraints as a useful diagnostic for studying learning dynamics, while showing that OrthoGrad is regime-dependent rather than universally regularizing.
\end{abstract}

\section{Introduction}
Neural networks can achieve near-perfect training accuracy even when trained on corrupted or randomly assigned labels, showing that high training accuracy alone does not imply meaningful generalization \cite{ZhangEtAl2017}. This phenomenon makes noisy-label learning a useful experimental setting for studying overfitting, memorization, and implicit regularization. Prior work has shown that deep networks often learn simple or structured patterns before eventually memorizing mislabeled examples, suggesting that noisy-label memorization is closely tied to training dynamics rather than model capacity alone \cite{ArpitEtAl2017,LiuEtAl2020}. Surveys on learning with noisy labels further emphasize that memorization of incorrect labels remains a central failure mode in supervised deep learning \cite{SongEtAl2020}.

A common way to control overfitting is to modify the objective function, architecture, or data distribution. Examples include explicit regularizers such as dropout, which randomly removes units during training to reduce co-adaptation, and weight decay, whose interaction with adaptive optimizers has motivated methods such as AdamW \cite{SrivastavaEtAl2014,LoshchilovHutter2019}. However, these interventions do not isolate the role of the optimizer’s update geometry. Since the choice of optimization algorithm itself can bias the solutions found by a neural network, studying update-level interventions provides another route for understanding implicit regularization \cite{NeyshaburEtAl2015,BlancEtAl2020}.

Geometric explanations of generalization must be handled carefully. Norms, sharpness, and other parameter-space quantities can change under rescaling or reparameterization without necessarily changing the function represented by the network. \citep{DinhEtAl2017} show that sharpness-based explanations can be misleading because equivalent networks can have very different apparent sharpness, while \citep{NeyshaburEtAl2015} argue that appropriate geometry should account for rescaling invariances in neural networks \cite{DinhEtAl2017,NeyshaburEtAl2015}. Therefore, this paper does not treat raw parameter-space geometry as a complete explanation of generalization. Instead, it uses geometry as an experimental intervention: we ask whether constraining the direction of optimizer updates changes the tendency of a model to memorize noisy labels.

Orthogonal gradient methods provide a natural framework for such an intervention. In continual learning, Orthogonal Gradient Descent projects new-task gradients into a subspace intended to preserve performance on previous tasks \cite{FarajtabarEtAl2020}. In multi-task learning, PCGrad modifies conflicting task gradients by projecting one gradient onto the normal plane of another, reducing destructive interference between tasks \cite{YuEtAl2020}. These methods suggest that changing gradient directions, without directly changing the model architecture or dataset, can meaningfully affect training behavior.

This work studies a single-task analogue of orthogonal gradient modification. Rather than projecting gradients against previous-task directions, we project each gradient away from the current weight vector. Related single-task work has recently studied orthogonalizing gradients with respect to weights for neural calibration, showing that such geometric optimizer modifications can affect confidence and uncertainty behavior \cite{Hedges2025}. Motivated by this line of work, we study whether a similar update constraint can reduce noisy-label memorization. Our hypothesis is that the current weight vector represents directions already reinforced by training; removing the component of the gradient parallel to this vector may discourage repeated amplification of those directions and encourage a more tangential trajectory through parameter space.

We evaluate this intervention in noisy-label image classification, focusing on small-data regimes where memorization is especially visible. Our results show that OrthoGrad can reduce corrupted-label fitting and improve test accuracy in the clearest MNIST CNN settings. However, the effect is not uniform: in larger-data regimes and in CIFAR-10 ResNet-18 experiments, OrthoGrad modifies the memorization trajectory but does not consistently improve final accuracy or prevent eventual memorization. Thus, we do not present OrthoGrad as a universal regularizer. Instead, we use it as a controlled geometric intervention showing that optimizer update direction can shape memorization dynamics.

\section{Method}
Let $g=\nabla_w L(w)$ be the current gradient for a weight tensor $w$. OrthoGrad removes the component of $g$ parallel to $w$:
\[
g_{\perp}
=
g - \lambda \frac{\langle g,w\rangle}{\|w\|^2+\epsilon}w .
\]
Unless otherwise stated, all main experiments use the fixed full-projection setting $\lambda=1$. We apply this transformation to weight tensors only and exclude biases and normalization parameters. In the renormalized variant, we rescale the projected gradient to preserve the original gradient norm:
\[
\tilde g_{\perp}
=
\frac{\|g\|}{\|g_{\perp}\|+\epsilon} g_{\perp}.
\]
Adam then uses $\tilde g_{\perp}$ rather than the raw gradient in its usual moment updates. This leaves the architecture, loss function, dataset, learning rate, and Adam hyperparameters unchanged, isolating the effect of update direction.

\section{Results}
We evaluate OrthoGrad on noisy-label image classification. Our main experiments use MNIST with two lightweight architectures: a small MLP and a small CNN. The MLP flattens the $28\times 28$ input and uses two ReLU hidden layers of sizes 64 and 32 before a 10-class output layer. The CNN uses four convolutional layers with ReLU activations, one max-pooling layer, and two final linear layers.

For the main noisy-label setting, we train on 500 MNIST examples with $30\%$ label noise for 400 epochs using Adam with learning rate $10^{-3}$ and batch size 64. We use five matched random seeds for Adam and Adam+OrthoGrad. We report test accuracy, training accuracy, and corrupted-label accuracy, defined as the fraction of intentionally corrupted examples classified according to their corrupted training label.

\begin{figure}[ht]
\centering
\begin{minipage}[t]{0.48\textwidth}
    \centering
    \includegraphics[width=\linewidth]{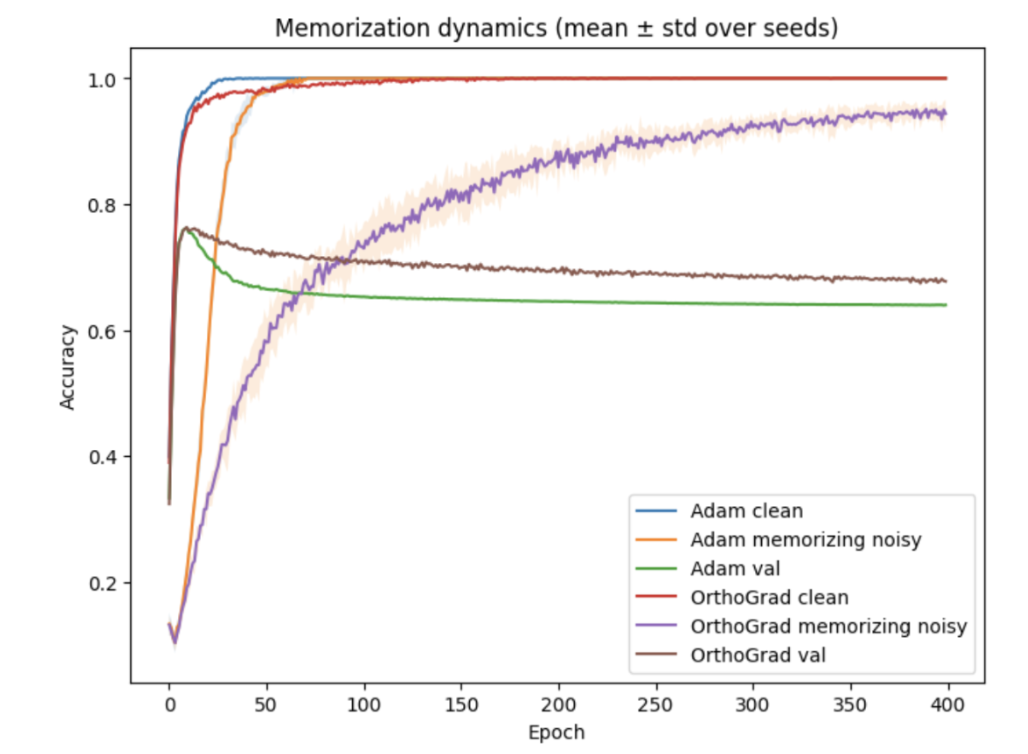}\\[-0.25em]
    {\scriptsize Small MLP}
\end{minipage}\hfill
\begin{minipage}[t]{0.48\textwidth}
    \centering
    \includegraphics[width=\linewidth]{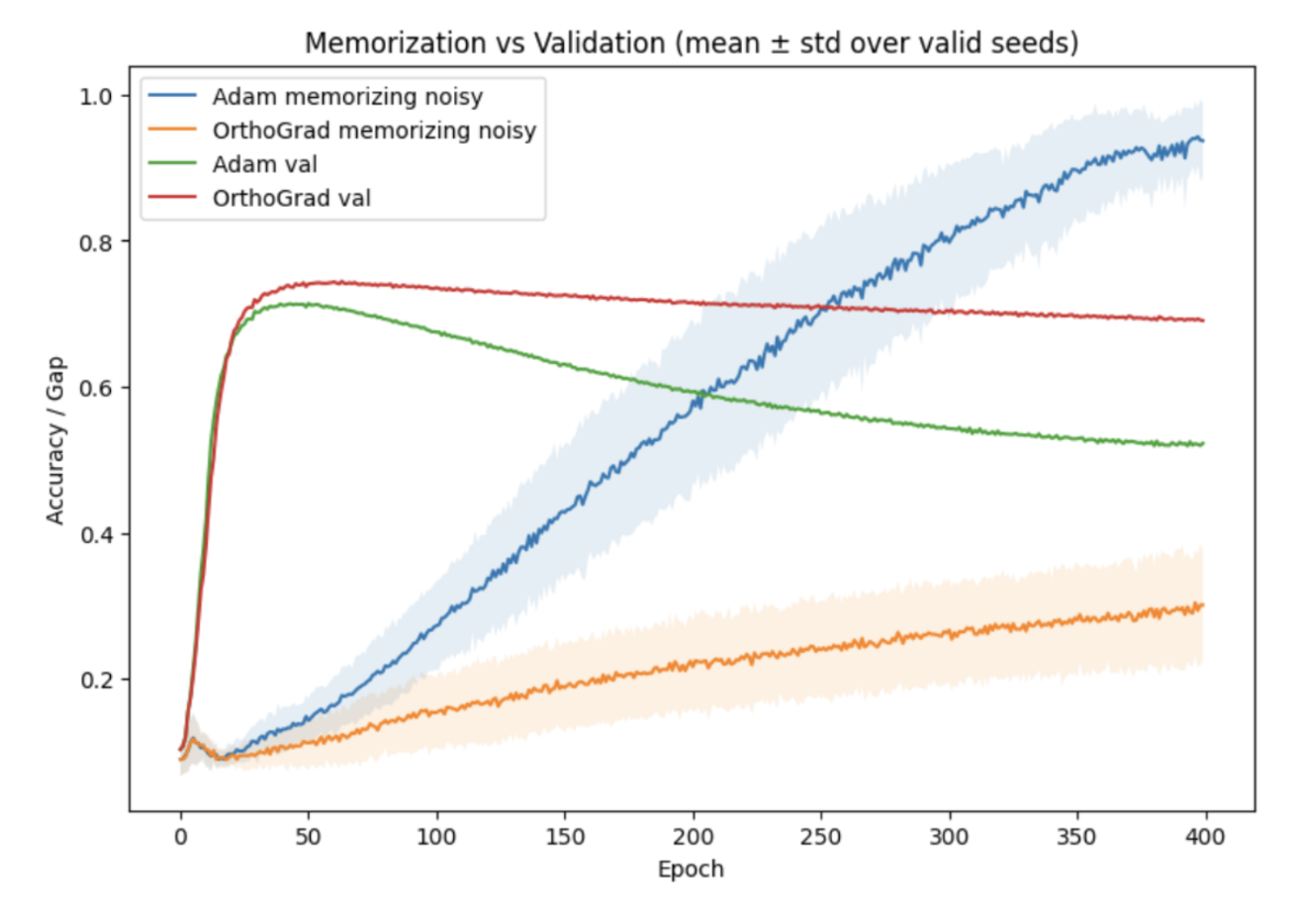}\\[-0.25em]
    {\scriptsize Small CNN}
\end{minipage}
\caption{Memorization and validation dynamics under 30\% noise.}
\label{fig:model-architectures}
\end{figure}

OrthoGrad most clearly helps in the noisy MNIST CNN setting. For the small MLP, test accuracy improves from $64.0\pm0.8\%$ under Adam to $67.7\pm0.7\%$ under OrthoGrad. For the small CNN, test accuracy increases from $52.2\pm2.5\%$ to $69.0\pm4.6\%$, while training accuracy decreases from $96.8\pm2.8\%$ to $71.2\pm3.1\%$. OrthoGrad also lowers corrupted-label accuracy, especially for the CNN, indicating that it suppresses fitting of the intentionally flipped labels.

The geometric intuition is that after projection, the update has no first-order radial component relative to the current weight vector. If $g_\perp \perp w$, then
\[
\|w-\eta g_\perp\|^2
=
\|w\|^2+\eta^2\|g_\perp\|^2,
\]
so the update is tangent to the weight-norm sphere to first order, although finite learning rates can still change the norm at second order. This motivates tracking weight norms and gradient-weight cosine similarity during training.

\textbf{Mechanism Analysis}.
To understand why OrthoGrad changes generalization behavior, we track the weight norm and the pre-projection cosine similarity between the gradient and the weight vector.

\begin{figure}[p]
\centering
{\small \textbf{Weight norm across training sizes}}\\[0.25em]
\begin{minipage}[t]{0.48\textwidth}
    \centering
    \includegraphics[width=0.94\linewidth]{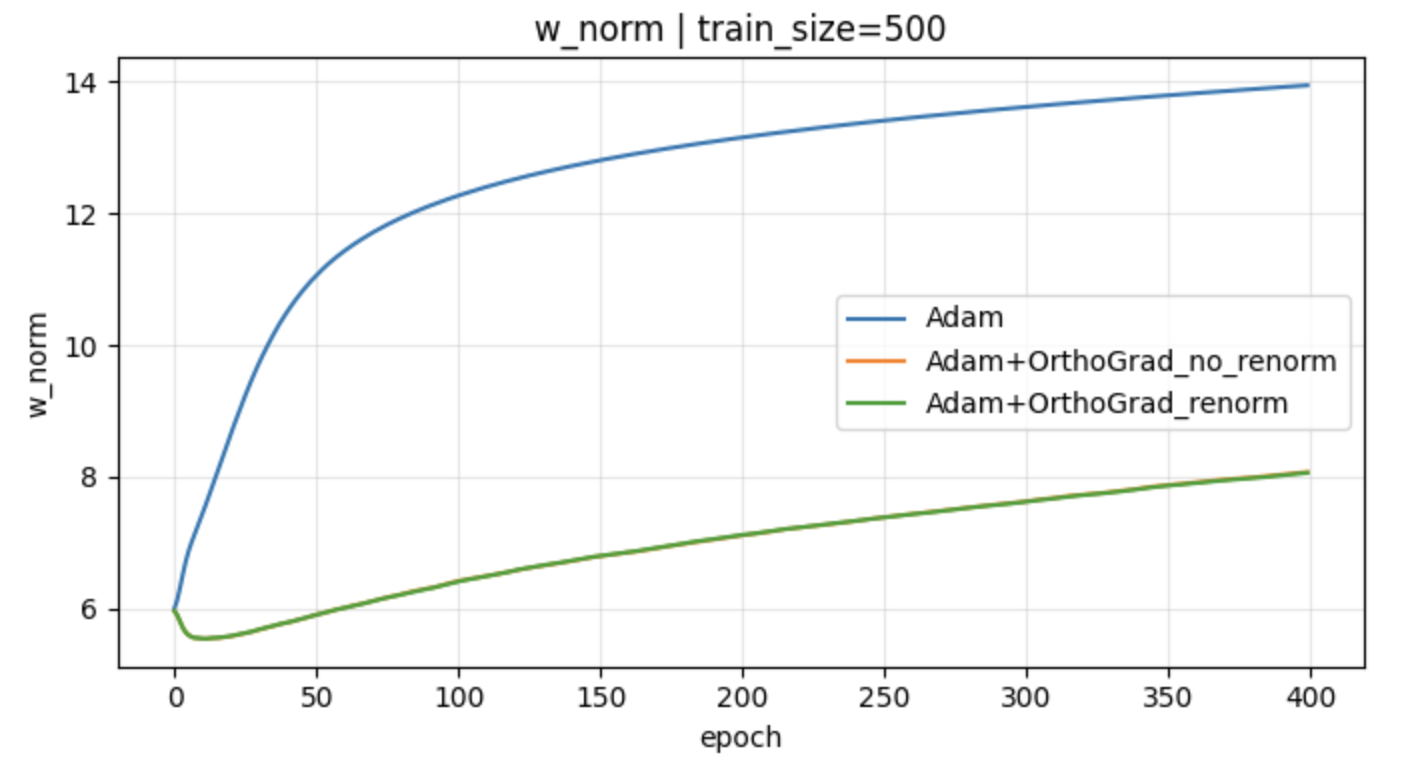}
\end{minipage}\hfill
\begin{minipage}[t]{0.48\textwidth}
    \centering
    \includegraphics[width=0.94\linewidth]{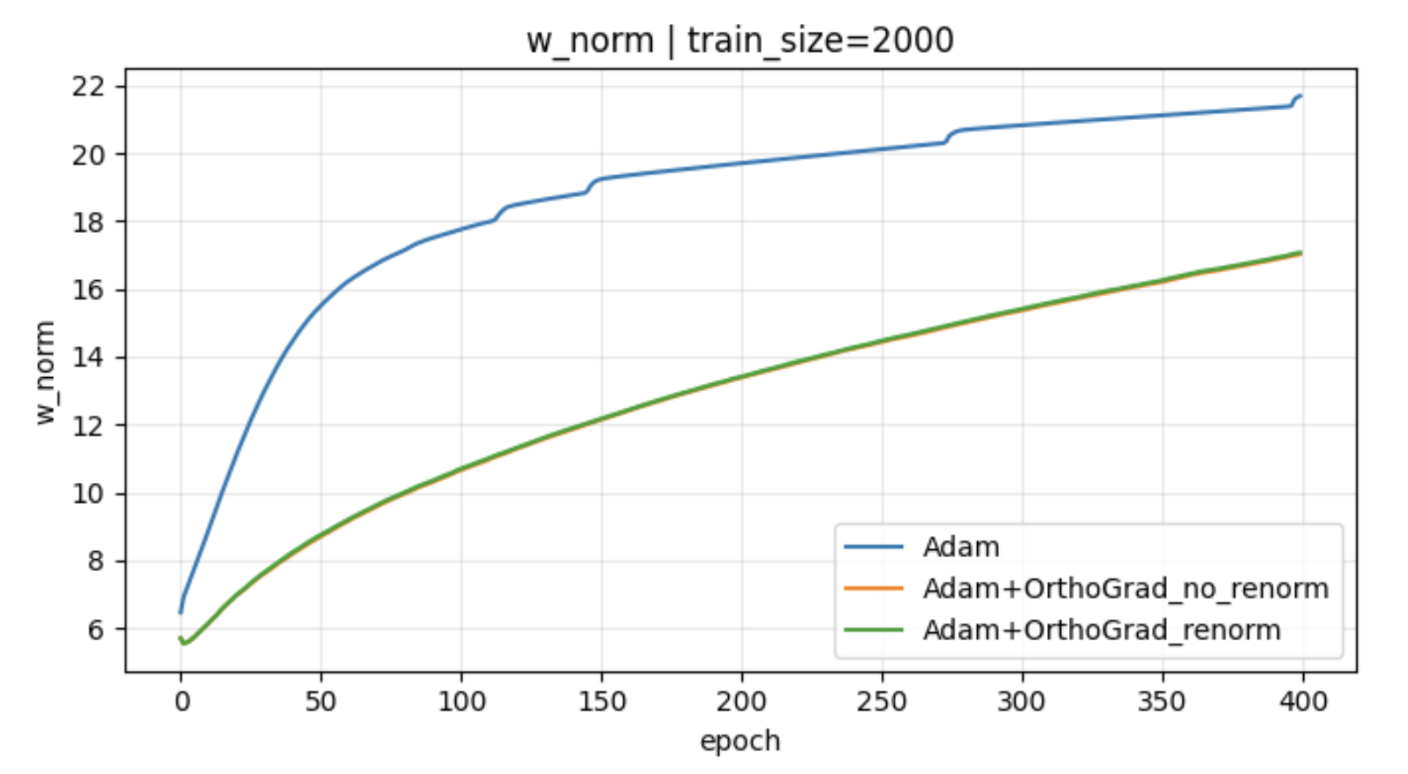}
\end{minipage}

\vspace{0.35em}

\begin{minipage}[t]{0.48\textwidth}
    \centering
    \includegraphics[width=0.94\linewidth]{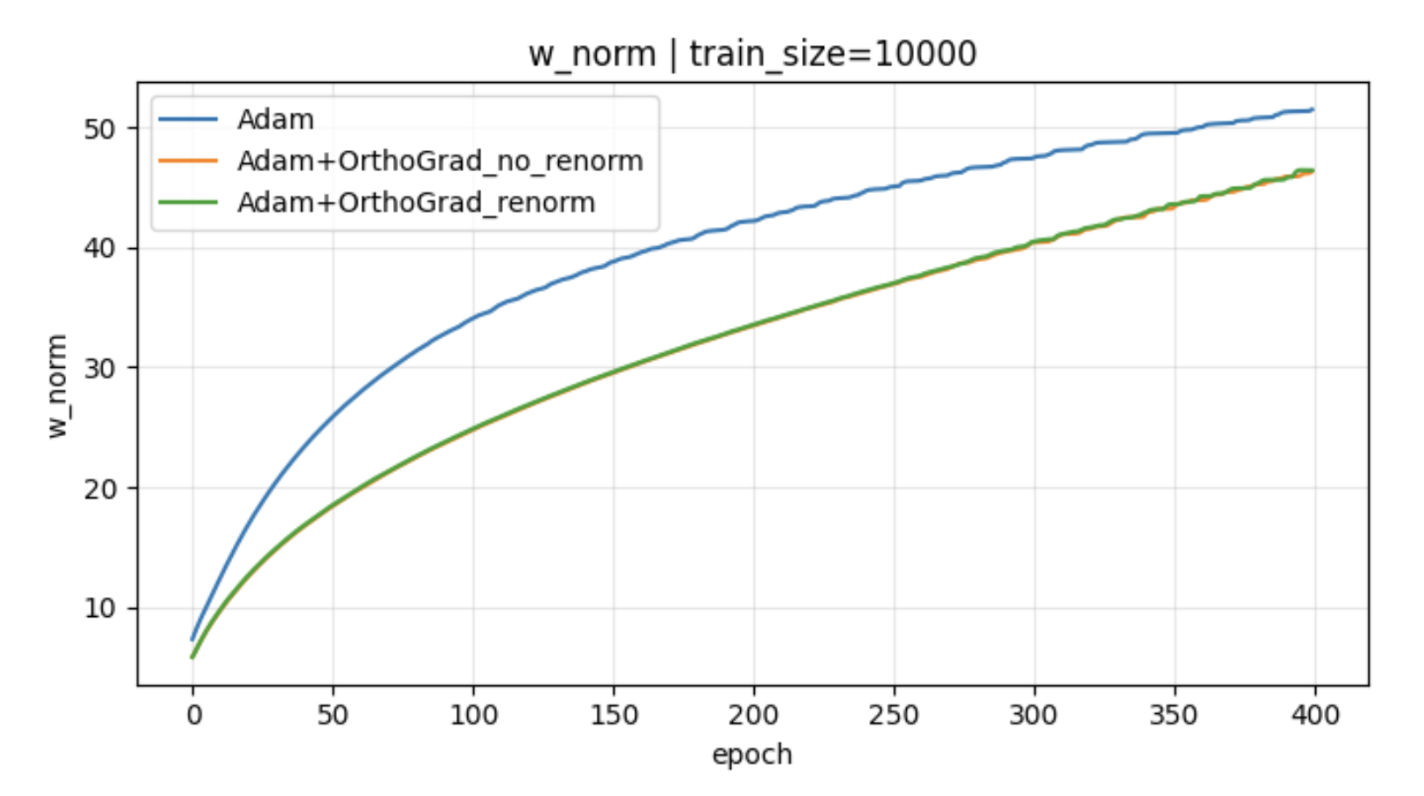}
\end{minipage}\hfill
\begin{minipage}[t]{0.48\textwidth}
    \centering
    \includegraphics[width=0.94\linewidth]{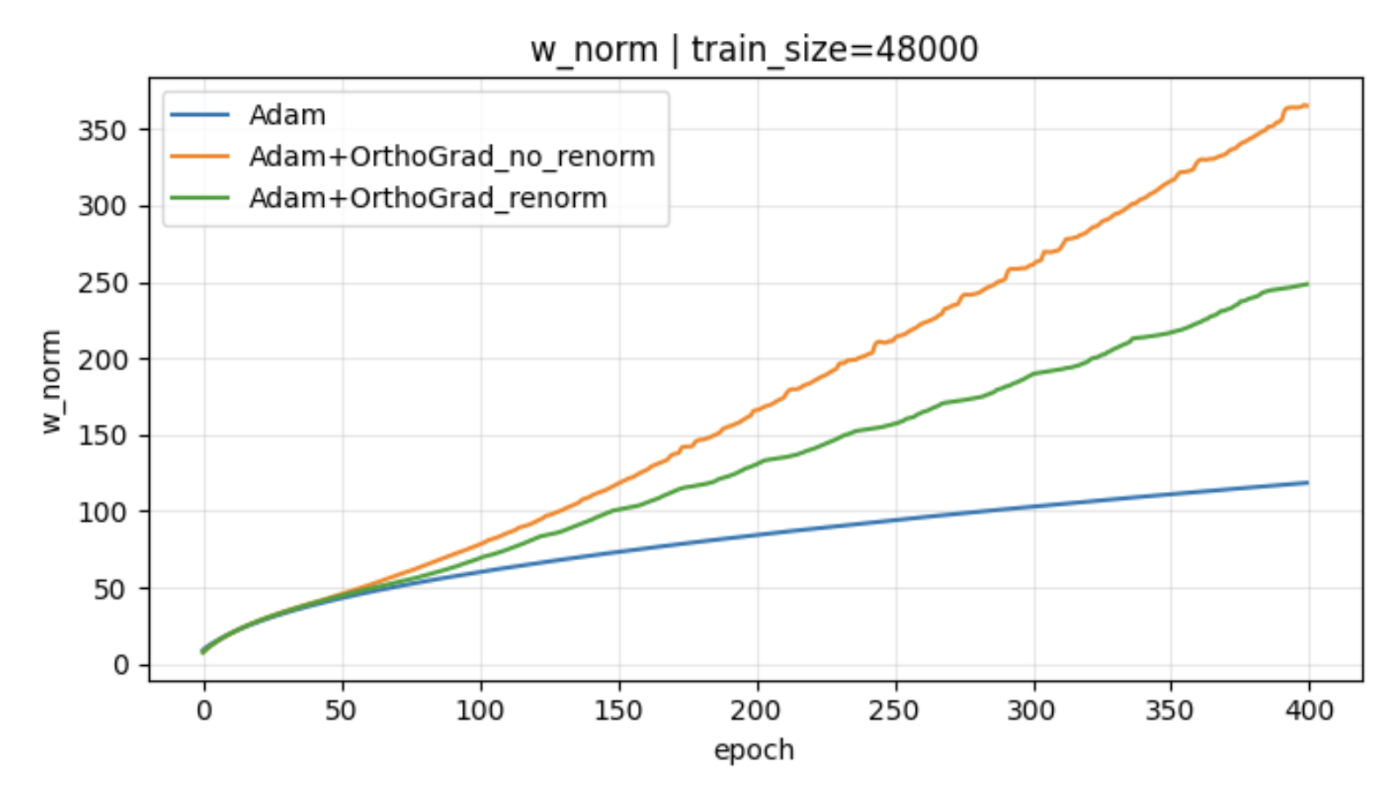}
\end{minipage}

\vspace{0.95em}

{\small \textbf{Pre-projection cosine similarity across training sizes}}\\[0.25em]
\begin{minipage}[t]{0.48\textwidth}
    \centering
    \includegraphics[width=0.94\linewidth]{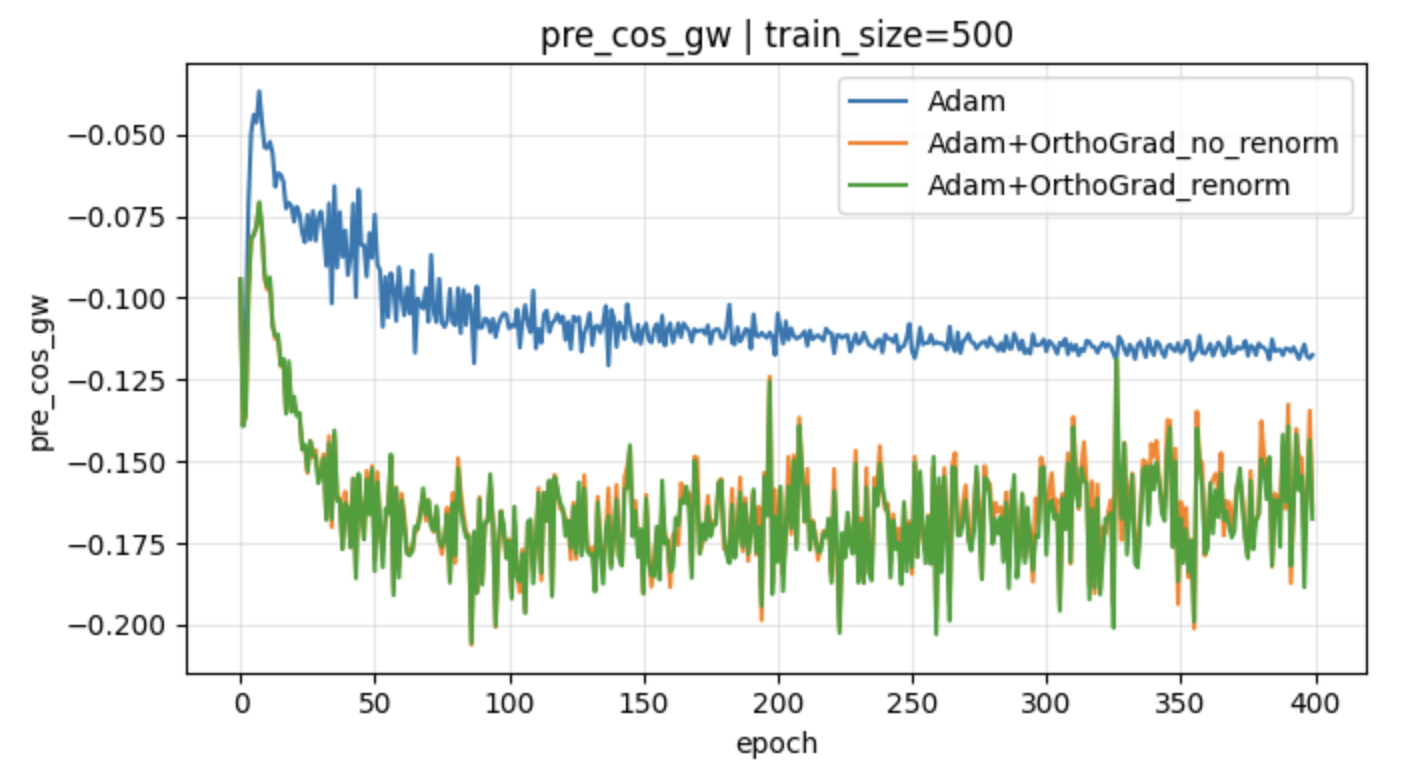}
\end{minipage}\hfill
\begin{minipage}[t]{0.48\textwidth}
    \centering
    \includegraphics[width=0.94\linewidth]{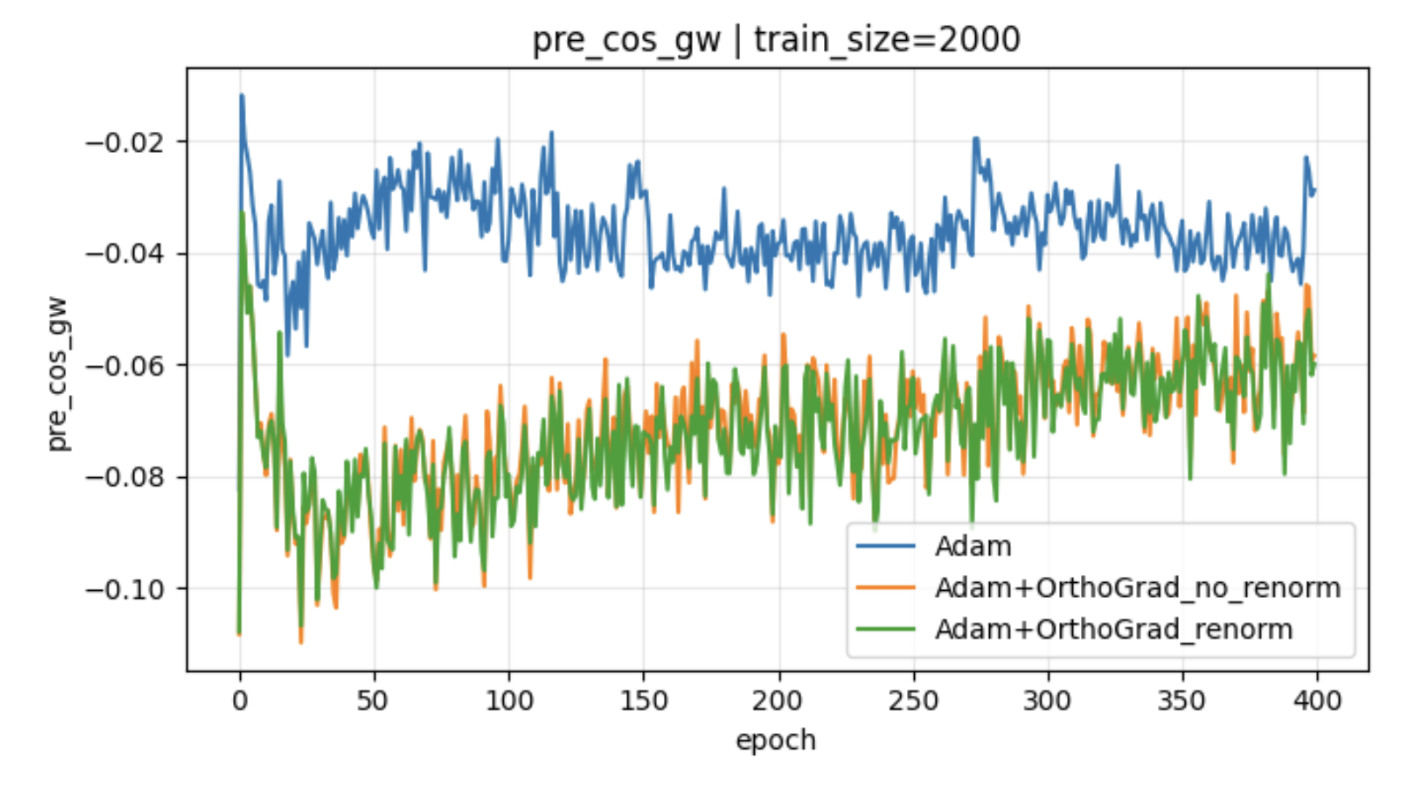}
\end{minipage}

\vspace{0.35em}

\begin{minipage}[t]{0.48\textwidth}
    \centering
    \includegraphics[width=0.94\linewidth]{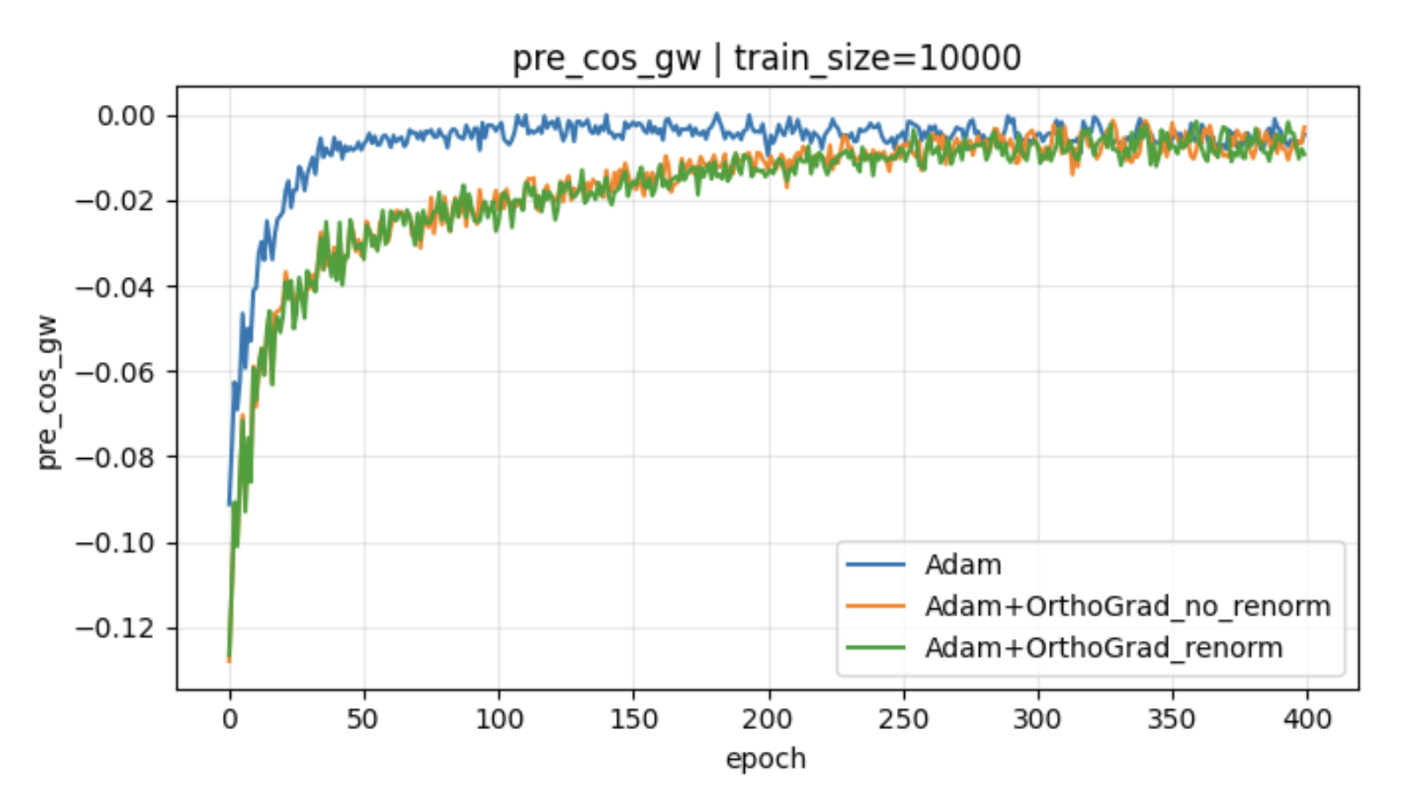}
\end{minipage}\hfill
\begin{minipage}[t]{0.48\textwidth}
    \centering
    \includegraphics[width=0.94\linewidth]{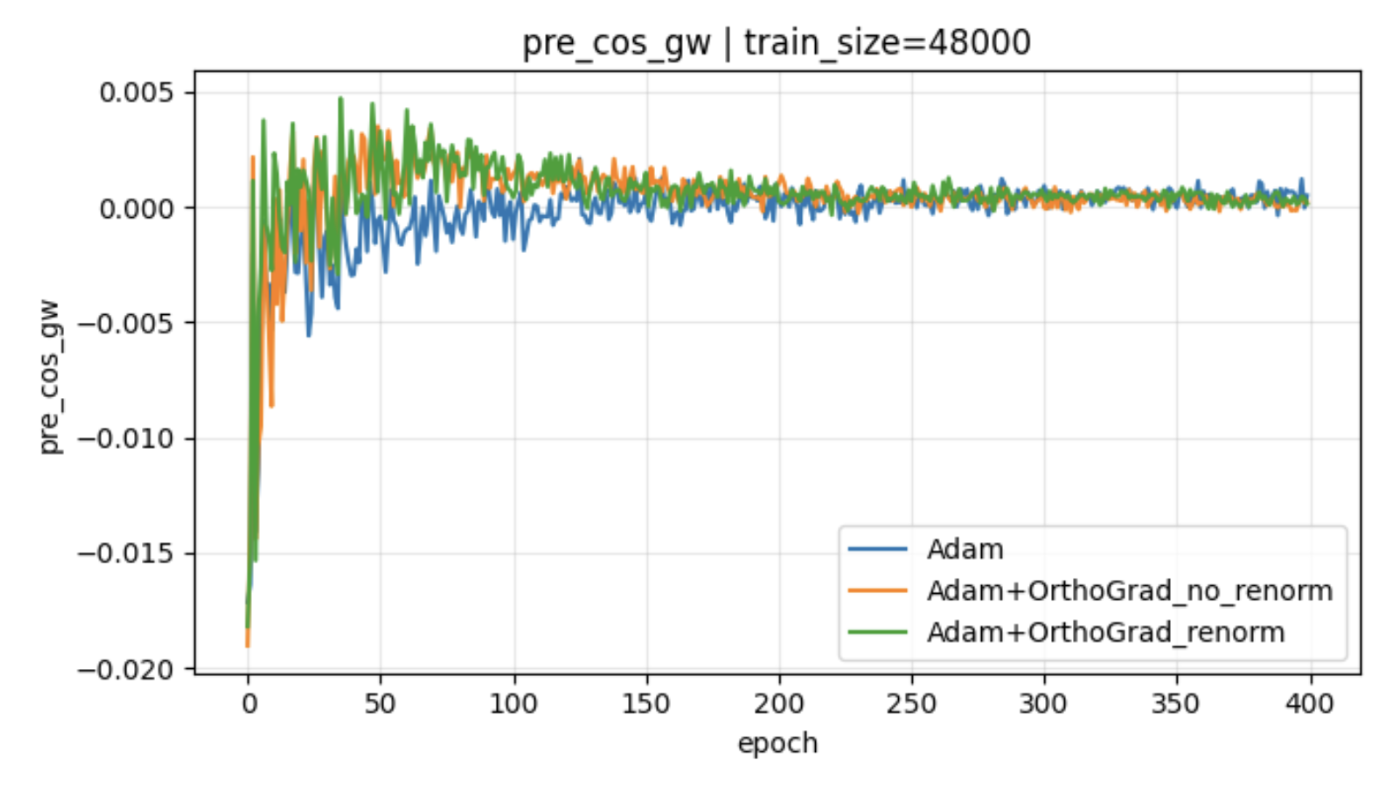}
\end{minipage}
\caption{Mechanism diagnostics across training sizes. In smaller-data settings, OrthoGrad changes the norm trajectory and the raw gradient has a nontrivial radial component. At full MNIST scale, gradient-weight cosine similarity is closer to zero and the projection has a weaker or qualitatively different effect, with OrthoGrad producing larger final norms.}
\label{fig:mechanism-group}
\end{figure}

\textbf{Mechanism Analysis Results.}
Figure~\ref{fig:mechanism-group} shows that the effect of OrthoGrad depends strongly on data regime. At $n=500$, OrthoGrad keeps the weight norm lower than Adam and the pre-projection gradient-weight cosine similarity is noticeably negative, so the projection removes a meaningful radial component of the update. At $n=48000$, the cosine similarity stays much closer to zero, meaning the raw gradient is already nearly orthogonal to the weight vector. In this regime, the projection is closer to a no-op directionally and no longer acts as a simple norm-stabilizing constraint. This helps explain why the strongest generalization effect appears in small-data noisy-label settings rather than full-data training.

\paragraph{CIFAR-10 extension.}
To test whether the MNIST trend transfers to a more complex image setting, we ran additional CIFAR-10 experiments using a ResNet-18 with 5000 training examples. On clean CIFAR-10, OrthoGrad was competitive with the baseline but did not show a consistent advantage. In the $20\%$ noisy-label setting, both methods eventually memorized corrupted labels, so final-epoch accuracy alone underestimates the trajectory-level effect. The baseline showed the expected memorization pattern: accuracy on corrupted labels increased over training while accuracy with respect to the original clean labels collapsed. OrthoGrad modified this trajectory in some runs, but did not prevent eventual memorization. We therefore interpret the CIFAR experiments as evidence that orthogonal update constraints can reshape memorization dynamics, not as evidence that OrthoGrad solves noisy-label learning in general.

\section{Discussion}
OrthoGrad resembles norm-based regularization in that it can constrain radial movement in weight space, but it is not equivalent to weight decay. Weight decay changes the objective or optimizer by explicitly shrinking weights, whereas OrthoGrad changes the direction of the gradient update. This distinction matters because OrthoGrad can reduce radial gradient components without directly optimizing for small parameter norm. Our experiments suggest that this update-level intervention is most effective when the raw gradient has a nontrivial component aligned with the current weight vector.

The gradient-weight cosine diagnostics also suggest why the effect weakens in larger regimes. In high-dimensional parameter spaces, unrelated directions are often close to orthogonal, but the relevant quantity here is not random-vector geometry alone; it is the empirical alignment between the training gradient and the current weights. When this cosine similarity is near zero, projecting away the weight-parallel component changes little. When it is noticeably nonzero, as in the small-data noisy-label runs, the same projection can substantially alter the optimizer trajectory.

This study has several limitations. The strongest positive results are on MNIST-scale noisy-label experiments, and the CIFAR-10 ResNet-18 extension is mixed rather than uniformly positive. In particular, OrthoGrad can alter the memorization trajectory on noisy CIFAR-10 but does not prevent eventual fitting of corrupted labels after long training. Thus, the method should not be interpreted as a universal regularizer. Instead, our results suggest a narrower conclusion: weight-orthogonal gradient constraints provide a useful intervention for studying how update geometry affects memorization dynamics. Future work should test larger architectures and datasets, compare systematically against weight decay and early stopping, and study adaptive projection strengths based on gradient-weight alignment.

\section*{Acknowledgment}
Claude and ChatGPT were used to assist with writing and grammar. However, the idea, experiments, and analysis were conducted solely by the author.
\nocite{*}
\bibliography{biblio}

\end{document}